\documentclass[a4paper]{report}
\usepackage[utf8]{inputenc}
\usepackage[T1]{fontenc}
\usepackage{RJournal}
\usepackage{amsmath,amssymb,array}
\usepackage{booktabs}

\usepackage{bm}
\usepackage{float}
\usepackage{csquotes}
\usepackage{yhmath}

\newcommand{\boldx}{\bm{x}}
\newcommand{\boldX}{\bm{X}}
\newcommand{\boldxS}{\bm{x}_{S}}
\newcommand{\boldxMinusS}{\bm{x}_{-S}}
\newcommand{\boldxMinusj}{\bm{x}_{-j}}
\newcommand{\boldxi}{\bm{x}^{(i)}}

\newcommand{\boldhS}{\bm{h}_{S}}

\begin{document}


\fancyfoot{}
\sectionhead{}

\begin{article}
\title{\pkg{fmeffects}: An R Package for Forward Marginal Effects}
\author{Holger Löwe\footnotemark[1], Christian A. Scholbeck\footnotemark[1], Christian Heumann, Bernd Bischl and Giuseppe Casalicchio}

\footnotetext[1]{H. Löwe and C.A. Scholbeck contributed equally.}
\maketitle

\abstract{

Forward marginal effects have recently been introduced as a versatile and effective model-agnostic interpretation method particularly suited for non-linear and non-parametric prediction models. They provide comprehensible model explanations of the form: if we change feature values by a pre-specified step size, what is the change in the predicted outcome? We present the R package \CRANpkg{fmeffects}, the first software implementation of the theory surrounding forward marginal effects. The relevant theoretical background, package functionality and handling, as well as the software design and options for future extensions are discussed in this paper.
}

\section{Introduction}

A growing number of disciplines are adopting black box machine learning (ML) models to make predictions, including medicine \citep{rajkomar_ml_medicine, boulesteix_ml_medicine}, psychology \citep{dwyer_psychology_ml}, economics \citep{mullainathan_econometrics_ml, athey_economics_ml}, or the earth sciences \citep{dueben_climate_ml}. Although one can often observe a superior predictive performance of black box models (such as neural networks, gradient boosting, random forests, or support vector machines) over intrinsically interpretable models (such as generalized linear or additive models), their lack of transparency or interpretability is considered a major drawback \citep{breiman_two_cultures}. This has been a major driver in the development of model-agnostic explanation techniques, which are often referred to by the umbrella terms of interpretable ML \citep{molnar_iml} or explainable artificial intelligence \citep{kamath_xai_book}.
\par
Marginal effects (MEs) \citep{williams_margins} have been a mainstay of model interpretations in many applied fields such as econometrics \citep{greene_econometric_analysis}, psychology \citep{mccabe_me_psychology}, or medical research \citep{onukwugha_me_primer}. MEs explain the effect of features on the predicted outcome in terms of derivatives w.r.t. a feature or forward differences in prediction. They are typically averaged to an average marginal effect (AME) for an entire data set, which serves as a global (scalar-valued) feature effect measure \citep{bartus_marginal_effects}. 
To explain feature effects for non-linear models, \citet{scholbeck_fme} introduced a unified definition of forward marginal effects (FMEs), a non-linearity measure (NLM) for FMEs, and the conditional average marginal effect (cAME).
The NLM is an auxiliary model diagnostic to avoid interpreting local changes in prediction as linear effects. The cAME aims to describe the model via regional FME averages for subgroups with similar FMEs, which can, for instance, be found by recursive partitioning (RP). FMEs, therefore, represent a means to explain models on a local, regional, and global level.
\par
\textbf{Contributions:} We present the R package \CRANpkg{fmeffects}, the first software implementation of the theory surrounding FMEs, including the NLM and the cAME. The user interface only requires a pre-trained model and an evaluation data set. The package is designed according to modular principles, making it simple to maintain and extend. This paper introduces the relevant theoretical background of FMEs, demonstrates the usage of the package in the context of a practical use case, and explains the software design.

\section{Background on forward marginal effects}

FMEs can be used for model explanations on the local, regional (also referred to as semi-global), and global level. These differ with respect to the region of the feature space that the explanation refers to. The local level explains a model/prediction for single observations, the regional level for a certain subspace (or subgroups of observations), and the global level for the entire feature space. 
Increasing the scope of the explanation requires increasing amounts of aggregations of local explanations (see the illustration by \citet{scholbeck_framework} of aggregations of local explanations to global ones for various methods). This can be problematic for non-parametric models where local explanations can be highly heterogeneous due to non-linear effects or interactions.

\subsection{Notation}

Let $\widehat{f}:\mathcal{X} \rightarrow \mathbb{R}$ be the prediction function of a learned model where $\mathcal{X} \subset \mathbb{R}^{p}$ denotes the feature space. While our definition naturally covers regression models, for classification models, we assume that $\widehat{f}$ returns the score or probability for a predefined class of interest. A subspace of the feature space is denoted by $\mathcal{X}_{[\;]} \subseteq \mathcal{X}$. 
The random feature vector is denoted by\footnote{Bold letters denote vectors.} $\boldX = (X_1, \dots, X_p)$. 
Observations are denoted by $\boldx = (x_1, \dots, x_p) \in \mathcal{X}$.
A set of feature indices is denoted by $S \subseteq \{1, \dots, p\}$. 
We often index (random) vectors as $\boldxS$ or $\boldX_S$.
We denote set complements by $-S = \{1, \, \dots\, , \, p\} \; \setminus \; S$. 
With slight abuse of notation, we represent the partitioning of a vector into two arbitrary but disjoint groups by $\boldx = (\boldxS, \boldxMinusS)$, regardless of the order of features. 
For a single feature of interest, the set $S$ is replaced by an integer index $j$. 
We usually assume an evaluation data set $\mathcal{D} = \left(\boldxi\right)_{i = 1}^n$, with $\boldxi \in \mathcal{X}$, which may consist of both training and test data.

\subsection{Forward marginal effects}
\label{sec:forward_marginal_effects}

The FME can be considered a basic, local unit of interpretation. Given an observation $\boldx$, it tells us how the prediction changes if we change a subset of feature values $\boldxS$ by a vector of step sizes $\boldhS$.
\begin{align*}
\text{FME}_{\boldx, \boldhS} &= \widehat{f}(\boldxS + \boldhS, \boldxMinusS) - \widehat{f}(\boldx) \quad \quad \text{for continuous features $\boldxS$}
\end{align*}
\citet{scholbeck_fme} introduced an observation-specific categorical FME whose definition is congruent with the FME for continuous features. The categorical FME corresponds to the change in prediction when replacing $x_j$ by the reference category $c_j$:
\begin{equation*}
\text{FME}_{\boldx, c_j} = \widehat{f}(c_j, \boldxMinusj) - \widehat{f}(\boldx) \quad \quad \text{for categorical $x_j$}
\end{equation*}
Note that this definition of a categorical ME differs from the one that is typically found in fields like econometrics \citep{williams_margins}, where we set $x_j$ to a reference category for all observations and then record the change in prediction resulting from changing the reference category to another category.
\par
Furthermore, it is common to globally average MEs to an average marginal effect (AME) to estimate the expected local effect. For FMEs, this corresponds to:
\begin{align}
        \;\text{AME}_{\mathcal{D}, \boldhS} &= \widehat{\mathbb{E}_{\boldX} \left[\text{FME}_{\boldX, \boldhS} \right]} \nonumber \\
        &= \frac{1}{n} \sum_{i = 1}^n \left[\widehat{f}\left(\boldxS^{(i)} + \boldhS, \boldxMinusS^{(i)}\right) - \widehat{f}\left(\boldxi\right) \right]
        \label{eq:ame}
\end{align}
Note that for categorical feature changes and observations where $x_j = c_j$, the FME equals 0. In the \pkg{fmeffects} package, the categorical AME only consists of observations whose observed feature value differs from the selected category. This approach is motivated by our goal to explain the effects of \textit{changing feature values} on the predicted outcome. For instance, in Fig. \ref{fig:categ_fme}, we demonstrate the effect of rainfall on the daily number of bike rentals in Washington D.C. by switching each non-rainy day's precipitation status to rainfall. Considering all observations, including rainy days, would obfuscate the interpretation we desire from our model. However, it is important to remember that every AME comprises a different set of points.

\subsection{Step size selection}

The selection of step sizes is determined by contextual and data-related considerations \citep{scholbeck_fme}. First, the FME allows us to investigate the model according to specific research questions. For instance, we might be interested in the effects of a specific change in a patient's body weight on the predicted individual disease risk. Often, we are interested in an interpretable or intuitive step size. In the case of body weight, typically expressed in kilograms, we could use a 1kg change (for instance, instead of 1g) as a natural increment. Without contextual information, we could use a unit change as a reasonable default; or dispersion-based measures such as one standard deviation, percentages of the interquartile range, or the mean/median absolute deviation.

\subsection{Non-linearity measure}

\begin{figure}[H]
    \centering
    \includegraphics[width = 0.42\textwidth]{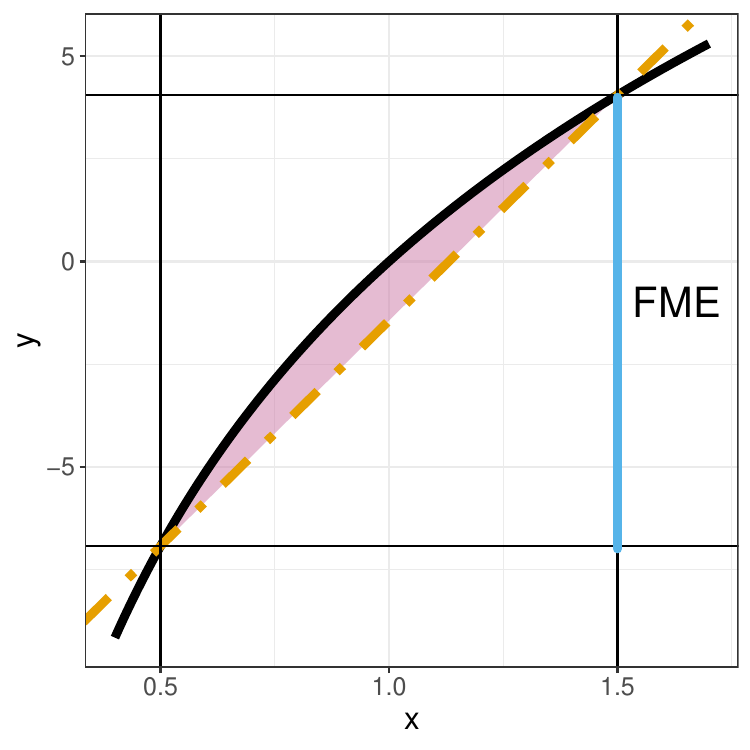}
    \caption{Illustration by \citet{scholbeck_fme} of a univariate FME (blue) given the prediction function (black) and linear secant (orange, dashed). The NLM indicates how well the secant can explain the prediction function (inversely proportional to the purple area) compared to how well the most uninformative baseline model (the average prediction) can explain the prediction function.}
    \label{fig:nlm_illustration}
\end{figure}

\begin{figure}[H]
    \centering
    \includegraphics[width = 0.49\textwidth]{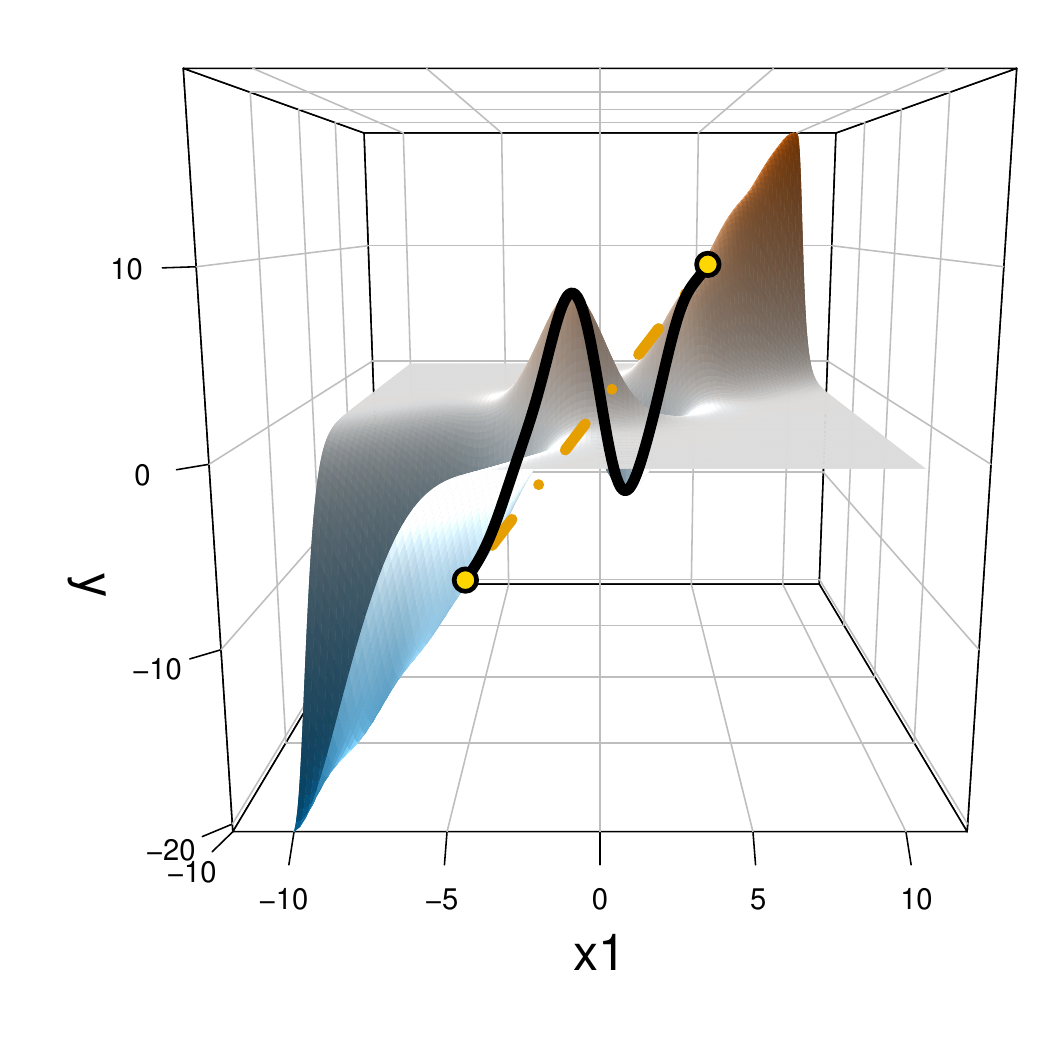}
    \includegraphics[width = 0.49\textwidth]{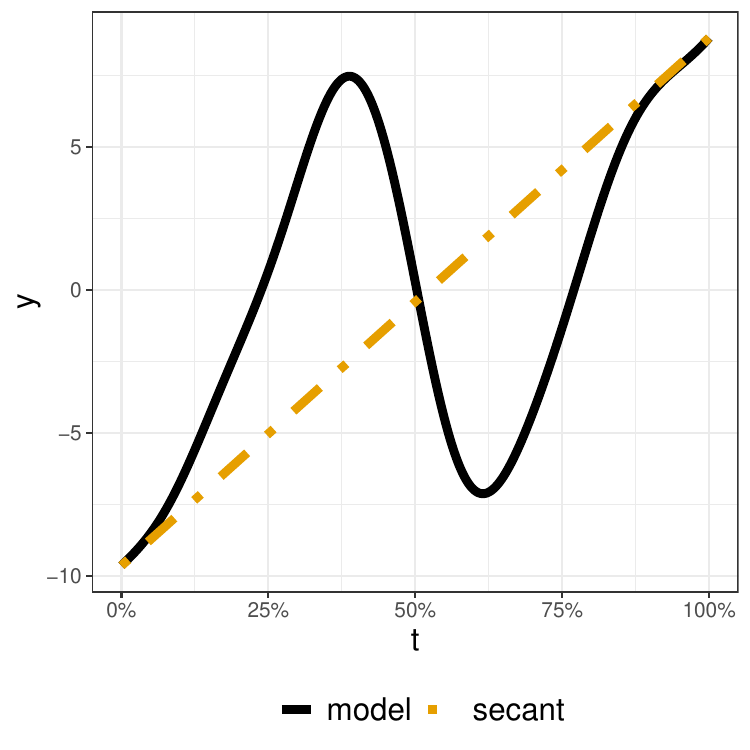}
    \caption{Illustration of the multivariate NLM by \citet{scholbeck_fme}. \textbf{Left:} An exemplary bivariate prediction function and two points to compute an FME. Consider an observation $\boldx = (-5, -5)$ and step size vector $\boldhS = (10, 10)$. We create the shortest path through the feature space to reach the point (5, 5), which consists of directly proportional changes in both features. Above the path, we see the linear secant (orange, dashed) and the non-linear prediction function (black). \textbf{Right:} The multivariate change in feature values can be parameterized as a percentage $t$ of the step size $\boldhS$. The deviation between the prediction function and the linear secant, as well as the deviation between the prediction function and mean prediction, both correspond to a line integral.}
    \label{fig:nlm_illustration_multivariate}
\end{figure}

For continuous features, we can consider $\boldx_S + \boldhS$ a continuous transition of feature values. The associated change in prediction may be misinterpreted as a linear effect. This is counteracted by the NLM, which corresponds to a continuous coefficient of determination $R^2$ between the prediction function and the linear secant that intersects $\boldx$ and $(\boldxS + \boldhS, \boldxMinusS)$ (see Fig. \ref{fig:nlm_illustration}). The continuous transition through the feature space is first parameterized as a fraction $t \in [0, 1]$ of the multivariate step size $\boldhS$:
\begin{align*}
    \gamma(t) = \begin{pmatrix} x_1 \\ \vdots \\ x_p \end{pmatrix} + t \cdot \begin{pmatrix} h_1 \\ \vdots \\ h_s \\ 0 \\ \vdots \\ 0  \end{pmatrix}
    \quad  , \quad t \in [0, 1] 
\end{align*}
The value of the linear secant $g_{\boldx, \boldhS}(t)$ corresponds to:
\begin{align*}
&\phantom{{}={}} g_{\boldx, \boldhS}(t) = \begin{pmatrix} x_1 + t \cdot h_1 \\ \vdots \\ x_s + t \cdot h_s \\ \vdots \\ x_p \\ \widehat{f}(\boldx) + t \, \cdot \, \text{FME}_{\boldx, \boldhS} \end{pmatrix} 
\end{align*}
The mean prediction $\widehat{f}_{\text{mean}}$ on the interval $t \in [0, 1]$ is given by: 
\begin{align*}
    \widehat{f}_{\text{mean}} &= \frac{\int_0^1 \widehat{f}(\gamma(t)) \; \Big\vert \Big\vert \frac{\partial \gamma(t)}{\partial t}\Big\vert \Big\vert_2 \;dt}{\int_0^1 \Big\vert \Big\vert \frac{\partial \gamma(t)}{\partial t} \Big\vert \Big\vert_2 \; dt}  \\
    &= \int_0^1 \widehat{f}(\gamma(t)) \;dt
\end{align*}
The NLM compares the squared deviation between the prediction function and the linear secant to the squared deviation between the prediction function and the mean prediction:
\begin{align*}
    \text{NLM}_{\bm{x}, \bm{h}_S} &= 1 - \frac{\int_0^1 \left(\widehat{f}(\gamma(t)) - g_{\boldx, \boldhS}(t)\right)^2 \; \Big\vert \Big\vert \frac{\partial \gamma(t)}{\partial t}\Big\vert \Big\vert_2 \;dt}{\int_0^1 \left(\widehat{f}(\gamma(t)) - \widehat{f}_{\text{mean}} \right)^2 \; \Big\vert \Big\vert \frac{\partial \gamma(t)}{\partial t}\Big\vert \Big\vert_2 \;dt} \quad
    \in (-\infty, 1]
\end{align*}
Fig. \ref{fig:nlm_illustration_multivariate} illustrates the setting for multivariate feature changes. The NLM can be approximated via numerical integration, e.g., via Simpson's rule.

The NLM indicates how well the linear secant can explain the prediction function, compared to the baseline model of using the mean prediction. A value of 1 indicates perfect linearity, where the linear secant is identical to the prediction function. For a value of 0, the mean prediction can explain the prediction function to the same degree as the secant. For negative values, the mean prediction better explains the prediction function than the linear secant (severe non-linearity).
\par
It is, therefore, easiest to interpret FMEs with NLM values close to 1. Although every FME always represents the exact change in prediction, an FME with a low NLM value does not fully describe the behavior of the model in that specific locality. In contrast, an FME with an NLM close to 1 is a sufficient descriptor of the (linear) model behavior. In other words, the NLM serves as an auxiliary diagnostic tool, indicating trust in how well the FME describes the local change in prediction.

\subsection{Conditional average marginal effect}

To receive a global model explanation akin to a beta coefficient in linear models, local FMEs can be averaged to the AME. \cite{mehrabi_survey_bias} define an \textit{aggregation bias} as drawing false conclusions about individuals from observing the entire population.
Given a data set $\mathcal{D}$, the conditional average marginal effect (cAME) estimator applies to a subgroup of $n_{[\;]}$ observations, denoted by $\mathcal{D}_{[\;]}$:

\begin{align}
    \text{cAME}_{\mathcal{D}_{[\;]}, \boldhS} &= \widehat{\mathbb{E}_{\boldX_{[\;]}} \left[\text{FME}_{\boldX_{[\;]}, \boldhS} \right]} \nonumber \\
    &= \frac{1}{n_{[\;]}} \sum_{i \,:\, \boldxi \in \mathcal{D}_{[\;]}} \left[\widehat{f}\left(\boldxS^{(i)} + \boldhS, \boldxMinusS^{(i)}\right) - \widehat{f}\left(\boldxi\right) \right] 
    \label{eq:came}
\end{align}

Although this estimator can be applied to arbitrary subgroups, we aim to find subgroups with cAMEs that counteract the aggregation bias.
Desiderata for such subgroups include within-group effect homogeneity, between-group effect heterogeneity, full segmentation, non-congruence, confidence, and stability \citep{scholbeck_fme}. In other words, we aim to partition the data into subgroups that explain variability in the FMEs. A viable option to partition $\mathcal{D}$ is to run RP on $\mathcal{D}$ with FMEs as the target. For instance, in \pkg{fmeffects}, both \CRANpkg{rpart} \citep{rpart} and \code{ctree()} from \CRANpkg{partykit} \citep{partykit} are supported to find subgroups.

\section{Related work}

\subsection{Model-agnostic interpretations}

The basic mechanism behind model-agnostic methods is to probe the model with different feature values, a methodology similar to a model sensitivity analysis \citep{scholbeck_framework, scholbeck_bridgingthegap}. The basis of explaining models is to determine the direction and magnitude of the effect of features on the predicted outcome \citep{casalicchio_featureimportance, scholbeck_framework, scholbeck_fme}. The individual conditional expectation (ICE) \citep{goldstein_ice}, partial dependence (PD) \citep{friedman_pdp}, accumulated local effects (ALE) \citep{apley_ale}, Shapley values \citep{strumbelj_shapley, lundberg_shap, covert_sage} and local interpretable model-agnostic explanations (LIME) \citep{ribeiro_lime} are some of the most popular model-agnostic explanation methods for ML models. Notably, counterfactual explanations \citep{wachter_counterfactuals} represent the reverse of the FME, indicating the smallest necessary change in feature values to reach a targeted prediction.
\par
FMEs complement the literature by allowing for a unique combination of local, regional, and global model explanations. Furthermore, while most methods (including the ICE, PD, ALE, or Shapley values) provide explanations in terms of prediction \textit{levels}, FMEs provide explanations in terms of prediction \textit{changes}. LIME is based on training a local and interpretable surrogate model whose coefficients can also provide an interpretation in terms of prediction changes. \citet{scholbeck_fme} highlighted differences between both approaches: notably, while surrogate models introduce additional uncertainty connected with the estimation of the surrogate, FMEs are motivated by the goal of stable and comprehensible model insight. Furthermore, locally estimated FMEs can be aggregated within subgroups and entire data sets for regional and global explanations. 
Around the same time, regional aggregations have also been introduced for ICE curves, for example \citep{britton_vine, herbinger_repid, molnar_cpfi}.

\subsection{Relationship between individual conditional expectation and forward marginal effect}

\citet{scholbeck_fme} illustrated a relationship between the ICE / PD and the FME / AME. In general, the FME can be seen as the difference between two locations on an ICE. The AME corresponds to the difference between two locations on the PD only for a function that is linear in the feature of interest. Therefore, the following relationship between the ICE and FME is worth noting here. The ICE can be considered a one-way sensitivity function that indicates the effects of varying a set of features indexed by $S$ while the remaining ones are kept constant:
\begin{equation*}
     \text{ICE}_{\boldx, S}(\boldx^{\ast}_S) = \widehat{f}(\boldx^{\ast}_S, \boldxMinusS)
\end{equation*}
For an instance $\boldx$, the prediction after increasing $\boldxS$ by $\boldhS$ is also a value of the ICE:
\begin{align*}
     \text{FME}_{\boldx, \boldhS} &= \widehat{f}(\boldxS + \boldhS, \boldxMinusS) - \widehat{f}(\boldx) \\
     &= \text{ICE}_{\boldx, S}(\boldxS + \boldhS) - \text{ICE}_{\boldx, S}(\boldxS)
\end{align*}

\subsection{Related work on marginal effects}

MEs have a long history in applied statistics and the Stata programming language \citep{stata_manual}. Initially implemented by \citet{bartus_marginal_effects}, the \code{margins()} command is now fully integrated into Stata and provides comprehensive capabilities for various computations and visualizations of statistical models such as (generalized) linear models \citep{williams_margins}. MEs are typically defined in terms of derivatives of the model w.r.t. a feature. For instance, this variant is the default approach to interpret models in econometrics \citep{greene_econometric_analysis}. The FME is the less commonly used definition \citep{scholbeck_fme, mize_discrete_change}. Note that\textemdash in contrast to forward differences\textemdash derivatives are not suitable to explain piecewise constant prediction functions such as tree-based models. 
\par
In recent years, MEs have gained traction in the R community. The R package \CRANpkg{margins} \citep{leeper_margins} was the first port of Stata's \code{margins()} command to R. Other packages related to MEs include \CRANpkg{ggeffects} \citep{ggeffects} and \CRANpkg{marginaleffects} \citep{marginaleffects}. In particular, \pkg{marginaleffects} can also return FMEs (although under different terminology). Our package, \pkg{fmeffects}, mainly differs from \pkg{marginaleffects} in two aspects:
\begin{description}
    \item[Implementing new theory surrounding FMEs:] The \pkg{fmeffects} package is the first software implementation of the theory surrounding model-agnostic FMEs as introduced by \citet{scholbeck_fme}. Although packages such as \pkg{marginaleffects} support the computation of FMEs and other quantities, \pkg{fmeffects} is specifically designed for FMEs with unique features such as implementations of the NLM, the cAME via RP, and novel visualization methods.
    \item[Model-agnostic black box interpretations:] 
    It follows that \pkg{fmeffects} is targeted at model-agnostic explanations of non-linear or intransparent models. Whereas existing theory on MEs (and packages such as \pkg{marginaleffects}) focuses on classical statistical modeling in combination with statistical inference (see, for instance, \citet{breiman_two_cultures} comparing statistical modeling culture with ML), FMEs (and thus \pkg{fmeffects}) are comparable to methods and software from the literature on interpretable ML such as the ICE, PD, ALE, or LIME. This does not imply that \pkg{marginaleffects} cannot be used for black box interpretations. As mentioned in the previous point, it also supports the computation of FMEs, e.g., in combination with \CRANpkg{mlr3}, but the focus of \pkg{fmeffects} lies on the interpretation of black box models through a specialized and targeted range of novel capabilities.
\end{description}

\section{Advantages and limitations of forward marginal effects}

\subsection{Advantages}

\begin{figure}[H]
    \centering
    \includegraphics[width = \textwidth]{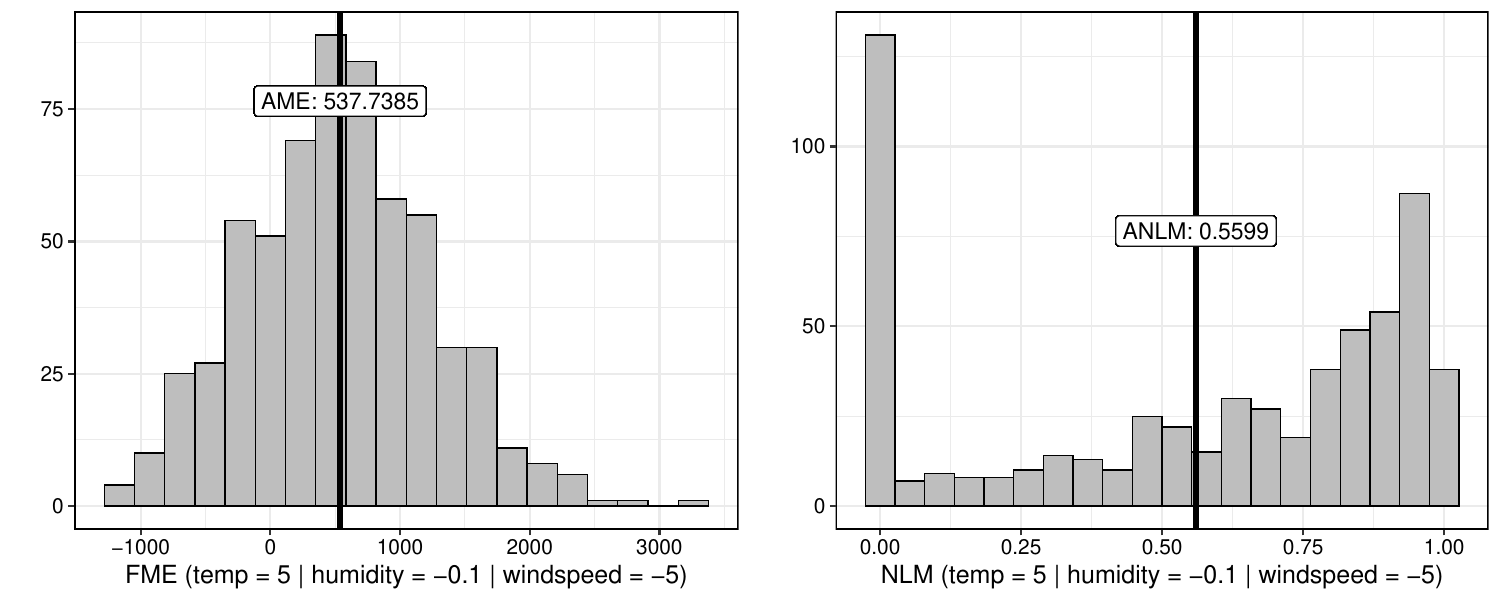}
    \caption{Adding a third feature change, a drop in  windspeed by 5 miles per hour, and visualizing the distribution of FME and NLM values. For the NLM plot, negative NLMs are binned as 0. It follows that the ANLM value in the plot differs from the raw ANLM in the summary output.}
    \label{fig:trivariate_change}
\end{figure}

Although the ICE and the FME are closely related, the latter provides several novel ways to generate insights into the model:

\begin{itemize}
    \item \textbf{Univariate changes in feature values:} FMEs are comparable to ICE curves for univariate changes in feature values. In certain scenarios, however, they may provide more comprehensible visualizations of effects for individual instances (see Fig. \ref{fig:iceplot} for an example). 
    \item \textbf{Bivariate changes in feature values:} The ICE and PD also provide insight into the sensitivity of the model prediction for variations in two features, which is visualized as a heatmap (see Fig. \ref{fig:bivariate_pdp}). However, it is difficult to visually compare the ICE of many different observations (which correspond to heatmaps as well). Although the ICE provides insight into a larger variation in feature values, while the FME only considers a single tuple of changes in feature values, bivariate FMEs can be easily compared visually (see Fig. \ref{fig:bivariate_fme}).
    \item \textbf{Higher-order changes in feature values:} If we evaluate the sensitivity of the prediction for changes in more than two feature values, virtually every visualization method breaks down. In this case, FMEs still provide comprehensible model explanations that can be aggregated in various ways (see Fig. \ref{fig:trivariate_change}).
    \item \textbf{Local fidelity assessment:}
    The locally restricted change in feature values for the FME facilitates evaluations of the fidelity of the model explanation (e.g., via the NLM). In other words, the NLM allows us to describe how well the FME summarizes the local shape of the prediction function in a single value. See Fig. \ref{fig:fme_nlm} for a visualization of NLM values for different observations.
    \item \textbf{Comprehensible regional explanations:} Although regional explanations have been first proposed in the context of grouping ICE curves \citep{herbinger_repid, britton_vine}, they more easily apply to scalar model explanations such as FMEs. Essentially, a regional model explanation represents a group of observations or a subspace of the feature space where model explanations are relatively homogeneous. Such groupings are easily achievable via RP or other techniques that do not require functional target values such as ICEs.
    \item \textbf{Avoiding extrapolation:} The ICE is computed on the entire feature range (see, e.g., Fig. \ref{fig:iceplot}), which is likely to result in model extrapolations. By its nature, the FME is typically used with small step sizes relative to the feature range, which naturally avoids model extrapolations.
\end{itemize}

\subsection{Limitations}

\begin{itemize}
    \item \textbf{Step size selection:} The step size fundamentally influences effects and the model interpretation. Although FMEs for different step sizes can be computed and visualized in an exploratory manner, some level of prior reasoning about what step sizes to use is recommended.
    \item \textbf{Decision tree instability for cAME:} Although not a shortcoming of the FME itself, subgroups found by RP to compute cAMEs are subject to a high variance. This may be counteracted by stabilizing the split search, e.g., by considering statistical significance of tree splits or resorting to different algorithms to find subgroups.
    \item \textbf{Non-linearity assessment for proportional feature changes:} For multi-dimensional feature changes, the NLM only considers equally proportional changes in all features.
\end{itemize}

\section{On causal interpretations and avoiding model extrapolations}

\label{sec:causality_extrapolations}

Note that model-agnostic techniques, including FMEs, explain associations between the target and the features within the model. In the absence of additional assumptions, such associations cannot be interpreted as causes and effects \citep{molnar_pitfalls}. For instance, increasing the value of a feature $x_1$ may always be accompanied by an increase in the target, but it may be the target $y$ that causes $x_1$ to increase. Another typical scenario is the presence of confounding factors that influence both $y$ and $x_1$. Finally, $x_1$ may only (or also) influence a mediator $x_2$, which in turn influences $y$.
\par
\begin{figure}[H]
    \centering
    \includegraphics[width = 0.65\textwidth]{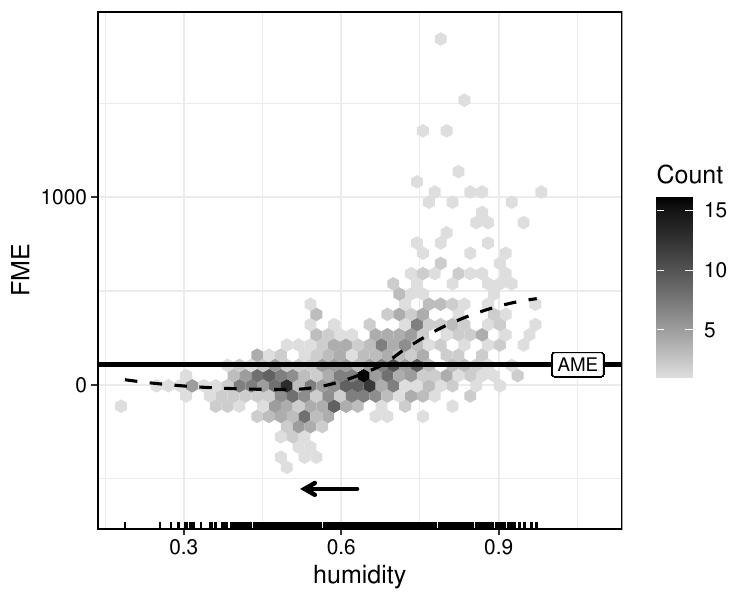}
    \caption{Univariate FMEs for a drop in humidity by 10 percentage points. Especially for high humidity values, the drop results in a considerable increase in predicted daily bike rentals.}
    \label{fig:univariate_fme_plot_humidity}
\end{figure}

This does not, however, make model interpretations obsolete. More importantly, as highlighted by \cite{Adadi}, model interpretations can be used to gain knowledge, debug, audit, or justify the model and its predictions. Throughout this paper, we will model the effects of environmental influences on the number of daily bike rentals in Washington, D.C. For our estimated model, a drop in humidity by 10 percentage points has a considerable effect on the predicted number of daily bike rentals (see Fig. \ref{fig:univariate_fme_plot_humidity}). This effect cannot be assumed to be causal, as humidity is physically influenced by the outside temperature, which will also affect people's choice to rent a bike. Here, temperature is a confounder that influences both humidity and daily bike rentals. However, the business renting out bikes can still use the associations found by a model with a good predictive performance to control the optimal number of bikes at their disposal. This is conditional on the model's ability to accurately predict the target for the given feature vector, requiring us to avoid model extrapolations, which correspond to predictions within areas of the feature space where the model has not seen much or any training data. This issue is closely linked to the multivariate distribution of the training data; in our example, a change in humidity is likely to be accompanied by a change in temperature as well, which we somewhat circumvent (depending on the magnitude of the step size) when making isolated changes to humidity. One may disregard this issue and deliberately predict in areas of the feature space the model has not seen during training. The resulting FMEs will still be valid model descriptions but, as explained above, they are likely to be bad descriptions of the data generating process.
\par
Model extrapolations negatively impact many model-agnostic interpretation methods \citep{hooker_fanova, hooker_cert, hooker_generalizedfanova, hooker_importance, molnar_pitfalls}. For example, \cite{apley_ale} demonstrated how PD plots suffer from extrapolation issues and introduced ALE plots as a solution to this problem. \citet{scholbeck_fme} illustrated the perils of model extrapolations for FMEs specifically and discussed possible options. One option in particular is also implemented in \pkg{fmeffects}: points outside the multivariate envelope (meaning the Cartesian product of all observed feature ranges) of the training data can be excluded from the analysis. This directly relates to the selection of small step sizes relative to the feature range, as large step sizes will result in a point falling outside the envelope.
\par
When using extrapolation prevention methods, note that we consider different sets of points for different step sizes, which differs from the usage of MEs in other contexts (see, for instance, the package \pkg{marginaleffects} for a comparison). The exclusion of points only impacts aggregations of FMEs, i.e., the cAME and AME. As discussed in the section on \nameref{sec:forward_marginal_effects}, this also affects the computation of categorical AMEs. In Eq. (\ref{eq:ame}) and Eq. (\ref{eq:came}), the AME and cAME are formulated as estimators of the expected global or regional (concerning a subspace) effects. The fewer observations we are considering for an average, the larger the variance of the estimate.

\section{User interface and package handling}
\label{sec:user_interface_package_handling}
\subsection{Local explanations}

The \code{fme()} function is the central user interface. 
It mainly requires a pre-trained model and a data set (see section \nameref{sec:design} for details). 
Further control parameters include a list of features and step sizes, whether to compute NLM values for each FME, and an extrapolation detection method. The \code{fme()} function initiates the construction and computations of a \code{ForwardMarginalEffect} object without requiring the user to know \CRANpkg{R6} \citep{r6} syntax.
\par
For this use case, we train a random forest from the \CRANpkg{ranger} package \citep{wright_ranger} on the bike sharing data set \citep{misc_bike_sharing_dataset_275} using \pkg{mlr3}. Note that models trained via \CRANpkg{tidymodels} and \CRANpkg{caret} are also supported, as well as models trained via \code{lm()}, \code{glm()}, and \code{gam()}. We aim to predict and explain the daily bike rental demand in Washington, D.C., based on features such as the outside temperature, wind speed, or humidity. We first train the model:

\begin{example}
> library(fmeffects)
> data(bikes, package = "fmeffects")
> library(mlr3verse)
> forest = lrn("regr.ranger")
> task = as_task_regr(x = bikes, id = "bikes", target = "count")
> forest$train(task)
\end{example}

\begin{figure}[H]
    \centering
    \includegraphics[width = 0.65\textwidth]{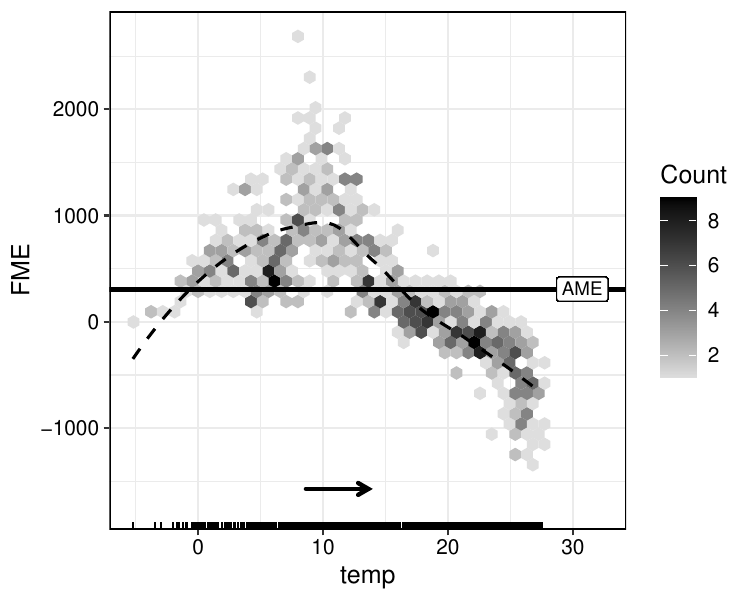}
    \caption{Plot of univariate FMEs for feature \samp{temp} and step size 5. Each hexagon represents a local FME average. The horizontal value represents the observed feature value of \samp{temp}. Each observation's \samp{temp} value is moved according to the arrow's direction and length. The vertical value of each hexagon indicates the FME value associated with that feature change. The horizontal bar indicates the AME. The shade of the hexagon implies how many observations it contains. A smoothing function facilitates interpretations by modeling an approximate pattern of FMEs across the feature range.}
    \label{fig:univariate_fme_plot_temp}
\end{figure}

Then, we simply pass the trained model, evaluation data, and remaining parameters to the \code{fme()} function. 
It returns a \code{ForwardMarginalEffect} object, which can be analyzed via \code{summary()} and visualized via \code{plot()} (see Fig. \ref{fig:univariate_fme_plot_temp}). Here, the outside temperature is raised by 5 degrees Celsius ceteris paribus. To avoid overplotting values, each hexagon represents a local average of FMEs. Users can easily access the data used by all plot functions to implement their own visualizations.
\par
Let us single out the observation with the largest associated FME. This observation corresponds to a single day with a recorded temperature of 8 degrees Celsius. Increasing the temperature by 5 degrees Celsius on this particular day results in 2699 additional predicted bike rentals. We plot such model explanations for the entire data set and average FMEs to receive a global model explanation. The AME\textemdash the global average of FMEs\textemdash is 307: an increase in temperature by 5 degrees Celsius results in an average increase of 307 predicted daily bike rentals.

\begin{example}    
> effects.univariate.temp = fme(
+   model = forest,
+   data = bikes,
+   features = list("temp" = 5),
+   ep.method = "envelope")

> summary(effects.univariate.temp)
\end{example}

\begin{example}
Forward Marginal Effects Object

Step type:
  numerical

Features & step lengths:
  temp, 5

Extrapolation point detection:
  envelope, EPs: 48 of 731 obs. (7 

Average Marginal Effect (AME):
  307.3275
 
> plot(effects.univariate.temp)
\end{example}
\par
\begin{figure}[H]
    \centering
    \includegraphics[width = 0.65\textwidth]{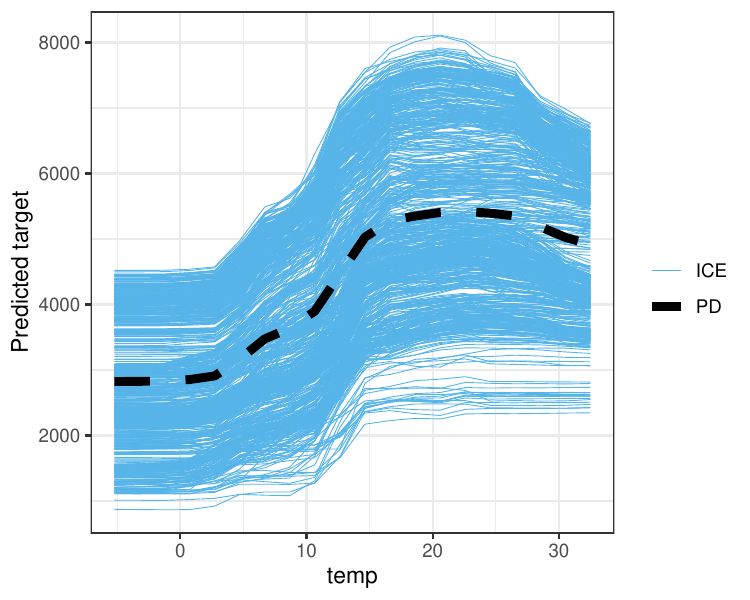}
    \caption{An ICE and PD plot for feature \samp{temp} generated by the R package \pkg{iml}. Each solid blue curve (an ICE) represents predictions for a single instance while only \samp{temp} varies. The dashed black curve (the PD) is the vertical average of ICEs and represents the average, isolated influence of \samp{temp}.}
    \label{fig:iceplot}
\end{figure}

Let us take a moment to compare the FME plot with the combined ICE and PD plot generated by the R package \CRANpkg{iml} \citep{molnar_imlpackage} (see Fig. \ref{fig:iceplot}).
This is one of the most popular and established model-agnostic ways to interpret predictive models \citep{molnar_iml}. The ICE is a local model explanation and represents the prediction for an observation where only the features of interest are varied (in this case, only \samp{temp}). The PD is the average of ICEs (in the univariate case, the vertical average) and indicates the global, average prediction when a subset of features is varied for all observations. 
Although we can see a rough trajectory of the feature influence on local and average predictions, it is difficult to pinpoint the exact effects of changing \samp{temp} on the prediction for single observations. Furthermore, ICE curves are more likely to be subject to model extrapolations, a result of predicting in areas where the model was not trained on a sufficient amount of data.

\par
FMEs allow for positive or negative step sizes. For instance, let us investigate the effects of an isolated drop in humidity by 10 percentage points. We can observe an AME of 108 additional predicted bike rentals a day. Individual effects tend to be larger the higher the humidity on that particular day.
\begin{example}    
> effects.univariate.humidity = fme(
+   model = forest,
+   data = bikes,
+   features = list("humidity" = -0.1),
+   ep.method = "envelope")

> summary(effects.univariate.humidity)
\end{example}

\begin{example}
Forward Marginal Effects Object

Step type:
  numerical

Features & step lengths:
  humidity, -0.1

Extrapolation point detection:
  envelope, EPs: 1 of 731 obs. (0 

Average Marginal Effect (AME):
  108.0477
\end{example}

\begin{figure}[H]
    \centering
    \includegraphics[width = 0.65\textwidth]{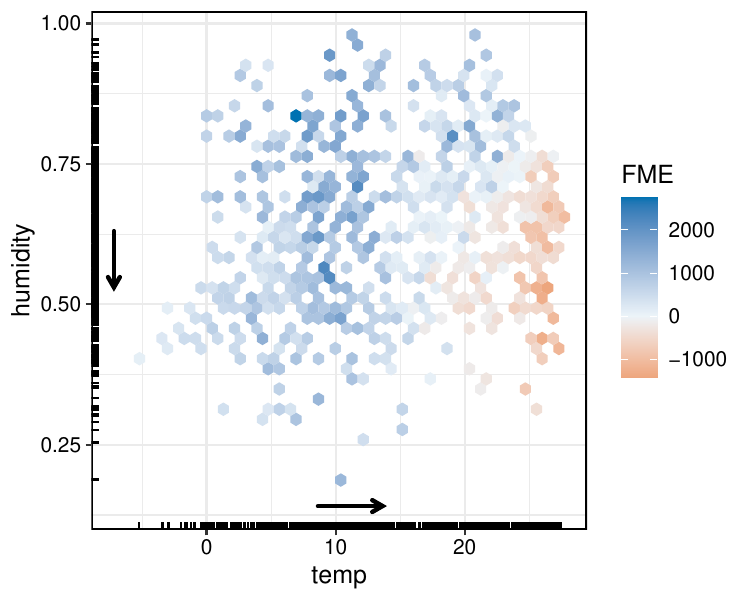}
    \caption{Visualizing bivariate FMEs for an increase in \samp{temp} by 5 degrees Celsius and a simultaneous drop in \samp{humidity} by 10 percentage points. FMEs are highly heterogeneous. We can see mostly positive effects, especially for observations with combinations of medium \samp{temp} and \samp{humidity} values.}
    \label{fig:bivariate_fme}
\end{figure}

\begin{example}    
> plot(effects.univariate.humidity)
\end{example}
\noindent In many applications, we are interested in interactions of features on the prediction. 
Until now, we only analyzed the univariate effects of \samp{temp} and \samp{humidity} on the predicted amount of bike rentals. However, potential interactions between features may exist. 
We evaluate an increase in temperature by 5 degrees Celsius and a simultaneous drop in humidity by 10 percentage points (see Fig. \ref{fig:bivariate_fme}). For a bivariate change in feature values, the two arrows depict the direction and magnitude of the feature change in the respective variable. As in the univariate case, we plot local averages within hexagons to avoid overplotting values. The location of the hexagon is determined by the observations' observed feature values in the provided data set. Its color indicates the FME associated with the bivariate feature change. An increase in the outside temperature by 5 degrees Celsius and a simultaneous drop in humidity by 10 percentage points is associated with an AME of 414. The univariate AMEs roughly add up to the bivariate AME, indicating that, on average, there is no additional interaction between both feature changes on the prediction.

\begin{example}
> effects.bivariate = fme(
+   model = forest,
+   data = bikes,
+   features = list("temp" = 5, "humidity" = -0.1),
+   ep.method = "envelope")

> summary(effects.bivariate)
\end{example}

\begin{example}
Forward Marginal Effects Object

Step type:
  numerical

Features & step lengths:
  temp, 5
  humidity, -0.1

Extrapolation point detection:
  envelope, EPs: 49 of 731 obs. (7 

Average Marginal Effect (AME):
  413.6163
\end{example}

\begin{figure}[H]
    \centering
    \includegraphics[width = 0.65\textwidth]{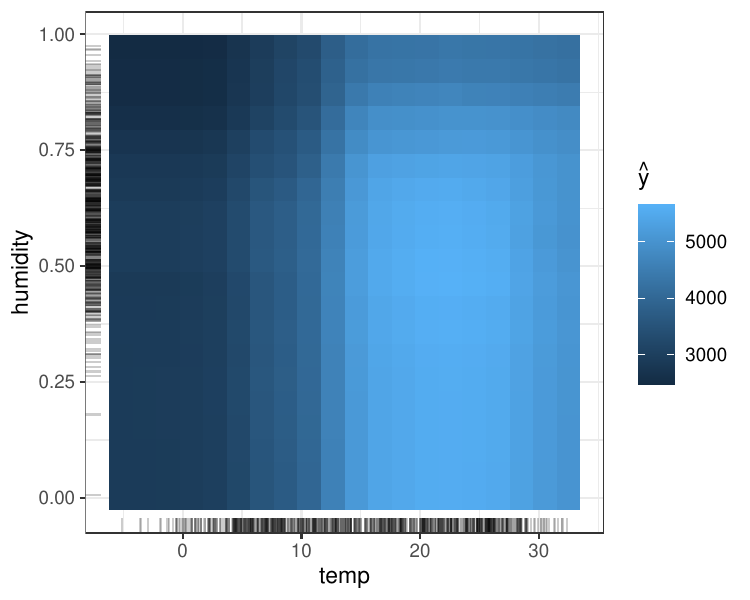}
    \caption{A bivariate PD plot (created via the R package \pkg{iml}), 
    visualizing the global interaction between \samp{temp} and \samp{humidity} on the predicted amount of bike rentals. Plugging in medium to large values for \samp{temp} and low to medium values for \samp{humidity}, ceteris paribus, results in more predicted bike rentals on average. 
    As opposed to bivariate FMEs, we cannot investigate multiple local effects, nor can we see the actual distribution of observed feature values. As a result, we cannot evaluate the effects of increasing one feature and decreasing another simultaneously.
    }
    \label{fig:bivariate_pdp}
\end{figure}

\begin{example}
> plot(effects.bivariate)
\end{example}
Let us repeat the same procedure as for univariate feature changes and compare the FME plot to an alternative option, the bivariate PD plot (see Fig. \ref{fig:bivariate_pdp}). 
As opposed to the novel visualization with FMEs, the PD plot only visualizes the average, global effect of changing both features on the predicted amount of bike rentals. It does not inform us about the distribution of observed feature values, thus not allowing us to evaluate the effects of increasing one feature and decreasing another simultaneously.

Let us now proceed to investigate non-linearity. Non-linearity can be visually assessed for ICE curves (see Fig. \ref{fig:iceplot}), but it is hard to quantify and would be somewhat meaningless for a large variation in the feature of interest. Furthermore, for bivariate or higher-dimensional changes in feature values, we lose any option for visual diagnoses of non-linearity. In contrast, the NLM can be computed for FMEs with continuous step sizes, regardless of dimensionality. The average non-linearity measure (ANLM) is 0.36, indicating that the linear secant, on average, is a bad descriptor of the FME.

\begin{figure}[H]
    \centering
    \includegraphics[width = 0.65\textwidth]{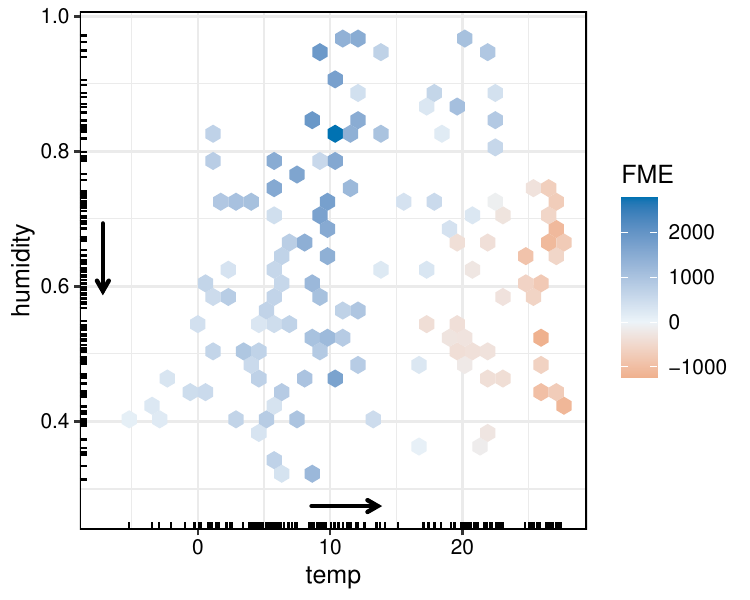}
    \caption{Visualizing FMEs with an NLM $\geq$ 0.9.}
    \label{fig:linear_fme_bivariate}
\end{figure}

\begin{example}
> effects.bivariate.nlm = fme(
+   model = forest,
+   data = bikes,
+   features = list("temp" = 5, "humidity" = -0.1),
+   ep.method = "envelope",
+   compute.nlm = TRUE)

> effects.bivariate.nlm

Forward Marginal Effects Object

Features & step lengths:
  temp, 5
  humidity, -0.1

Average Marginal Effect (AME):
  413.6163

Average Non-Linearity Measure (ANLM):
  0.36 

> plot(effects.bivariate.nlm, with.nlm = TRUE)
\end{example}
\begin{figure}[H]
    \centering
    \includegraphics[width = \textwidth]{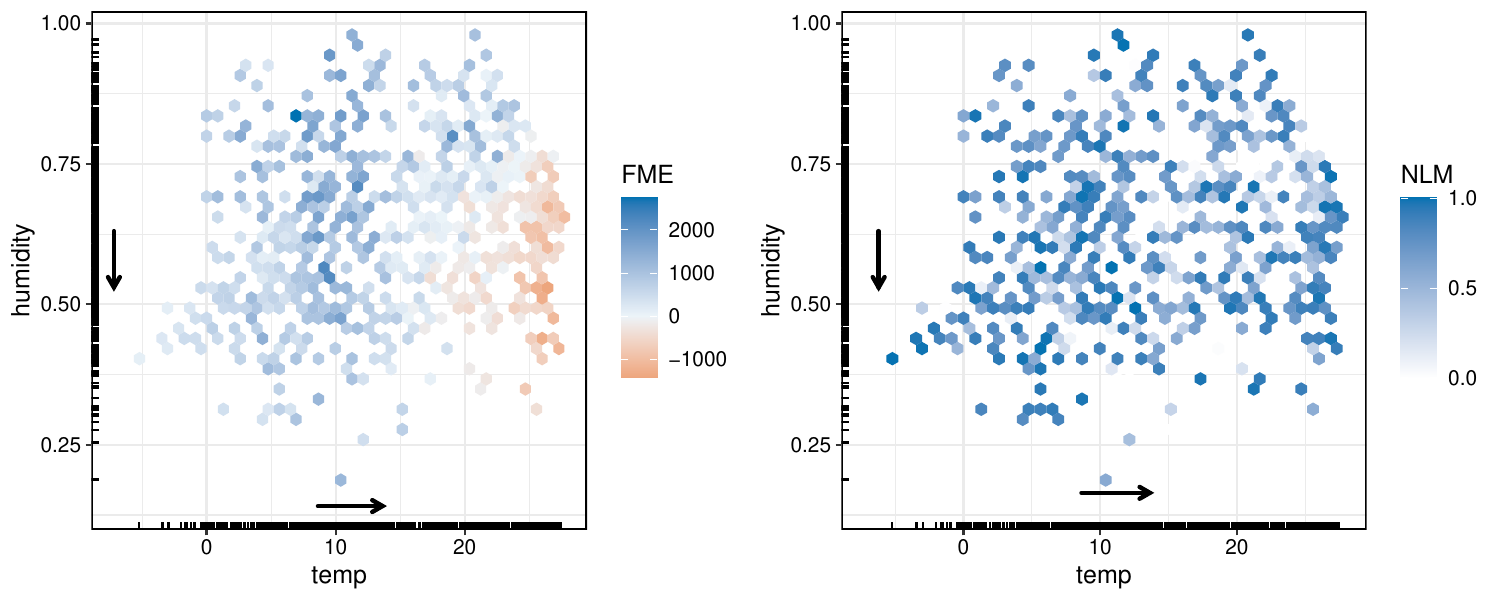}
    \caption{Adding NLM computations to the FME plot. Each hexagon in the left and right plots represents a local average of FME and NLM values, respectively. 
    }
    \label{fig:fme_nlm}
\end{figure}
Fig. \ref{fig:fme_nlm} simply contrasts FME values with the corresponding NLM values. In this case, we can see both non-linear FMEs (whiter NLM) and linear FMEs (deep blue-colored NLM). We could now, for instance, focus on interpreting linear FMEs. All FMEs depicted in Fig. \ref{fig:linear_fme_bivariate} have an NLM of 0.9 or higher, meaning that they almost fully describe the model prediction for proportional changes in \samp{temp} and \samp{humidity}.
\par
An advantage of FMEs is their ability to provide comprehensible model insight even when exploring higher-order feature changes. Let us factor in a third feature change, now simultaneously reducing windspeed by 5 miles per hour, and visualize the distribution of FME and NLM values. We can see that in addition to an increase in temperature and a decrease in humidity, a decrease in windspeed further boosts the average number of predicted daily bike rentals.
\begin{example}
> effects.trivariate.nlm = fme(
+   model = forest,
+   data = bikes,
+   features = list("temp" = 5, "humidity" = -0.1, "windspeed" = -5),
+   ep.method = "envelope",
+   compute.nlm = TRUE)

> summary(effects.trivariate.nlm)

Forward Marginal Effects Object

Step type:
  numerical

Features & step lengths:
  temp, 5
  humidity, -0.1
  windspeed, -5

Extrapolation point detection:
  envelope, EPs: 117 of 731 obs. (16 

Average Marginal Effect (AME):
  537.7385

Average Non-Linearity Measure (ANLM):
  0.33

> plot(effects.trivariate.nlm, with.nlm = TRUE)
\end{example}
\par
\begin{figure}[H]
    \centering
    \includegraphics[width = 0.5\textwidth]{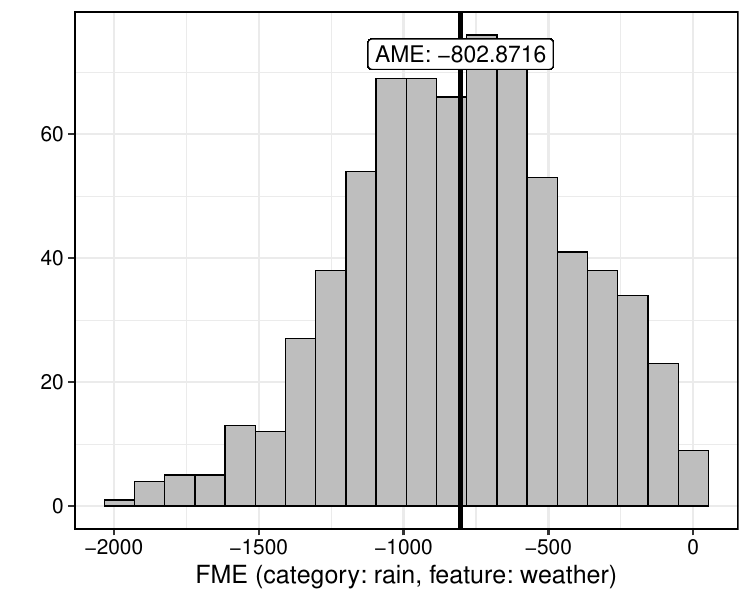}
    \caption{Distribution of categorical FMEs resulting from switching each non-rainy day's precipitation status to rain. On average, rainfall lowers predicted bike rentals by 803 bikes per day.}
    \label{fig:categ_fme}
\end{figure}

 So far, we have only evaluated changes in continuous features. In many applications, we are concerned with switching categories of categorical features, a way of counterfactual thinking inherent to the human thought process. Note that despite the allure of switching categories of categorical features, one needs to be aware of potential model extrapolations. To illustrate this, we switch each non-rainy day's precipitation status to rainfall. Rainfall has an average isolated effect of lowering daily rentals by 803 bikes (see Fig. \ref{fig:categ_fme}). 

\begin{example}
> effects.categ = fme(
+   model = forest,
+   data = bikes,
+   features = list("weather" = "rain"))

> summary(effects.categ)

Forward Marginal Effects Object

Step type:
  categorical

Feature & reference category:
  weather, rain

Extrapolation point detection:
  none, EPs: 0 of 710 obs. (0 

Average Marginal Effect (AME):
  -802.8716

> plot(effects.categ)
\end{example}

\subsection{Regional explanations}
\label{sec:regional_explanations}

In our examples, we can see highly heterogeneous local effects. The more heterogeneous FMEs are, the less information the AME carries. In many practical applications, we are interested in compactly describing the behavior of the predictive model across the feature space, akin to a beta coefficient in a linear model. This is where regional explanations come into play. We aim to find subgroups with more homogeneous FME values, thereby describing the behavior of the model not in terms of a global average but in terms of multiple regional averages (cAMEs).
\par
In \pkg{fmeffects}, this can be achieved by further processing the \code{ForwardMarginalEffect} object containing FMEs (and optionally NLM values) using the \code{came()} function. This returns a \code{Partitioning} object (in this case, an object of the class \code{"PartitioningCTREE"}, a subclass of the abstract class \code{"Partitioning"}, see later section on \nameref{sec:design}). 
\par
For the univariate change in temperature by 5 degrees Celsius, we decide to search for precisely 2 subgroups\footnote{This value is to be set by the user depending on how many regional explanations are to be found. Alternatively, we can search for a pre-defined SD of FMEs inside the terminal nodes. How many subgroups can be found depends on the data and predictive model.} (for a description of this algorithm, see the following section on \nameref{sec:design}).
A summary of the created object informs us about the number of observations, cAME, and standard deviation (SD) of FMEs inside the root node and leaf nodes (the found subgroups). We succeeded in finding subgroups with lower SDs while maintaining an appropriate sample size. The root node SD of 611 can be successfully split down to 437 and 355 within the subgroups. By visualizing the tree, we can see how the data was partitioned. For cooler outside temperatures equal to or lower than $\approx$ 16 degrees Celsius, we can observe a positive cAME of 728 additional bike rentals per day. On warmer days with a temperature above $\approx$ 16 degrees Celsius, the model predicts 196 less bike rentals a day when the outside temperature increases by 5 degrees.

\begin{example}
> subspaces = came(effects = effects.univariate.temp, number.partitions = 2)
> summary(subspaces)
\end{example}

\begin{example}
PartitioningCtree of an FME object

Method:  partitions = 2

   n      cAME  SD(fME)  
 683  307.3275 611.0778 *
 372  728.3942 437.0463  
 311 -196.3278 354.5090  
---
* root node (non-partitioned)

AME (Global): 307.3275
\end{example}

\begin{example}
> plot(subspaces)
\end{example}
\begin{figure}[H]
    \centering
    \includegraphics[width = \textwidth]{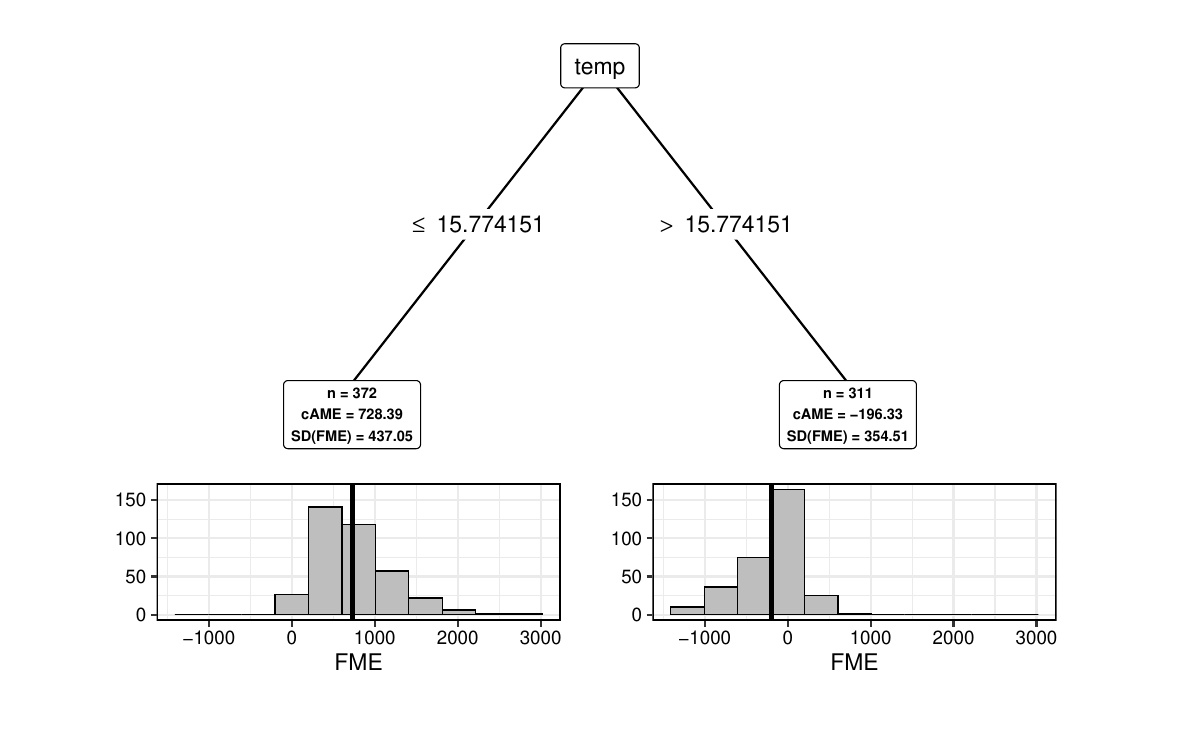}
    \caption{Using a decision tree to find subgroups of observations with more homogeneous FMEs of increasing \samp{temp} by 5 degrees Celsius. Each leaf node visualizes one subgroup, the number of observations, the cAME, and the SD of FMEs indicating FME homogeneity.}
\end{figure}

\subsection{Global explanations}

When to search for regional explanations thus depends on the heterogeneity of local effects. The \code{ame()} function provides an appropriate summary for the entire model.
It uses a default step size of 1 or 0.01 for small feature ranges. 
For categorical FMEs, it uses every observed category as a reference category. Alternatively, custom step sizes and subsets of features can be used.  
The \code{summary()} function prints a compact model summary of each feature, a default step size, the AME, the SD of FMEs, 25\% and 75\% quantiles of FMEs, as well as the number of observations left after excluding extrapolation points (EPs). A large dispersion indicates heterogeneity of FMEs and thus a small fidelity of the AME and possible benefits from searching for subgroups with varying cAMEs.
A different workflow can, therefore, also consist of starting with the table generated by \code{ame()} and deciding which feature effects can be described by AMEs and which might be better describable by subgroups and cAMEs. If this has been unsuccessful, we can resort to local model explanations. Recall our example from the previous section on \nameref{sec:regional_explanations} where we split FMEs associated with increasing temperature by 5 degrees Celsius. From the \code{ame()} summary, we see that \samp{temp} has a relatively large SD in relation to its AME (here calculated with a step size of 1), and the interquartile range indicates a wide spread of FMEs from -20 in the 25\% quantile up to 108 in the 75\% quantile, which makes it a promising candidate to find subgroups with more homogeneous FMEs.

\begin{example}
> ame.results = ame(model = forest, data = bikes)
> summary(ame.results)
\end{example}
\begin{example}
Model Summary Using Average Marginal Effects:

      Feature step.size        AME       SD       0.25       0.75   n
1      season    winter  -942.0906 466.3691 -1298.1011  -617.5663 550
2      season    spring   136.2185 569.5307  -244.4237   650.0125 547
3      season    summer   293.6264 549.2972   -42.7551   738.2056 543
4      season      fall   533.5502 579.5541    52.3706  1138.0863 553
5        year         0 -1899.4966 639.1695 -2354.1389 -1506.0582 366
6        year         1  1790.6269 524.4711  1421.7925  2194.1396 365
7     holiday        no     195.93  218.386   123.2468   228.0909  21
8     holiday       yes  -133.3134 154.8869  -201.3635   -25.1245 710
9     weekday    Sunday   155.5219 188.8708     9.3486   252.0308 626
10    weekday    Monday  -158.9218 215.5047  -263.2441    -4.8485 626
11    weekday   Tuesday  -115.7316 193.4508  -197.7396    13.3208 626
12    weekday Wednesday   -44.3056 173.8664  -115.5562    63.1344 627
13    weekday  Thursday     16.005  161.125   -61.1673    89.5043 627
14    weekday    Friday    57.1498 163.5602   -27.6519    128.752 627
15    weekday  Saturday   103.7648 170.5678    -0.2044    178.493 627
16 workingday        no   -42.8794 139.8572  -145.7104    66.2131 500
17 workingday       yes    48.1298 158.3666   -60.2448   145.5003 231
18    weather     misty  -221.5664 328.3458  -413.4363   -69.4238 484
19    weather     clear   385.8674 347.6119   162.2048   476.8631 268
20    weather      rain  -802.8716 384.2624 -1054.7158  -543.2614 710
21       temp         1    58.0487 164.8714   -20.0019   108.4669 731
22   humidity      0.01     -19.86  62.1753   -36.5407    10.4535 731
23  windspeed         1   -24.7315  77.1757   -56.9247    13.7468 731
\end{example}

\section{Design and options for extensions}
\label{sec:design}

\begin{figure}[htp]
    \centering
    \includegraphics[width=14cm]{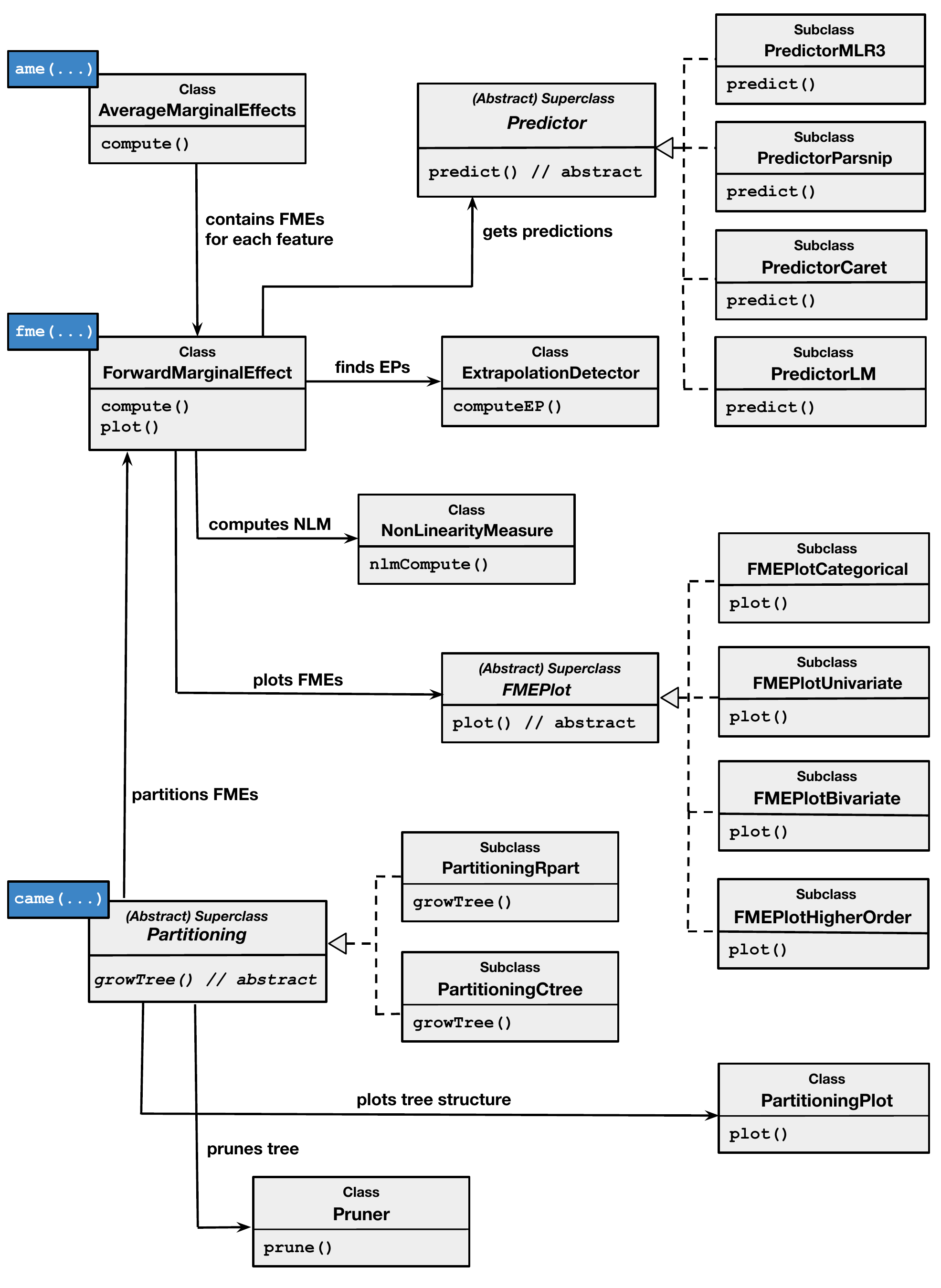}
    \caption{Design overview of the \pkg{fmeffects} package, including methods that implement the main functionality of each class. Classes may contain more methods than depicted.
    Blue boxes indicate wrapper functions to instantiate objects of the respective class.}
    \label{fig:architecture}
\end{figure}

The \pkg{fmeffects} package is built on a modular design for improved maintainability and future extensions. Fig. \ref{fig:architecture} provides a visual overview of the core design. The greatest emphasis is placed on the strategy and adapter design patterns \citep{gof_design}. Simply put, the strategy pattern decouples the source code for algorithm selection at runtime into separate classes. We repeatedly implement this pattern throughout the package by creating abstract classes whose subclasses implement specific functionalities. The adapter design pattern (also called a \enquote{wrapper}) creates an interface for communication between two classes.

\begin{itemize}
\item \code{"Predictor"}: An abstract class that implements the adapter pattern to accommodate future implementations of storing a predictive model. \code{"PredictorMLR3"}, \code{"PredictorParsnip"}, and \code{"PredictorCaret"} are subclasses that store an \pkg{mlr3}, \CRANpkg{parsnip} \citep{parsnip} (part of \pkg{tidymodels}), or \pkg{caret}  model object. This allows users of \pkg{fmeffects} to use numerous predictive models such as random forests, gradient boosting, support vector machines, or neural networks. \code{"PredictorLM"} stores models returned by \code{lm(), glm(), or gam()}. The package can be extended with novel model types by implementing a new subclass that stores the model, data, target, and is able to return predictions. 
\item \code{"AverageMarginalEffects"}: A class to compute AMEs for each feature in the data (or a subset of features). Internally, a new \code{"ForwardMarginalEffect"} object is used to compute and aggregate FMEs. For convenience, we implement a wrapper function \code{ame()} to facilitate object creation and to initiate computations without requiring user input in the form of \pkg{R6} syntax. 
\item \code{"ForwardMarginalEffect"}: The centerpiece class of the package. It keeps access to a \code{Predictor}, stores important information to create FMEs, and after the computations are completed, stores results and gives access to visualization methods. For convenience, the wrapper function \code{fme()} can be used.
\item \code{"FMEPlot"}: An abstract class for code decoupling of different plot categories into distinct classes. Subclasses include \code{"FMEPlotUnivariate"}, \code{"FMEPlotBivariate"}, \code{"FMEPlotHigherOrder"}, \\\code{"FMEPlotCategorical"}.
\item \code{"ExtrapolationDetector"}:
Identifies (and excludes) EPs. The current implementation supports the method \enquote{envelope}, excluding points outside the multivariate envelope of the training data.
\item \code{"NonLinearityMeasure"}: For the NLM, we need to approximate three line integrals, e.g., via Simpson's 3/8 rule. The general definition of Simpson's 3/8 rule for a univariate function f(x) and integration interval $[a, b]$ corresponds to:
\begin{equation}
\label{simpson}
\int_{a}^{b} f(x) \approx \frac{b-a}{8}\left[ f(a) + 3f\left(\frac{2a+b}{3}\right) + 3f\left(\frac{a+2b}{3}\right) + f(b) \right]
\end{equation}
We make use of a composite Simpson rule, which divides up the interval $\left[ a, b \right]$ into $n$ subintervals of equal size and approximates each subinterval with Eq. (\ref{simpson}).
\par
\item \code{"Partitioning":} An abstract class, allowing for various implementations of finding subgroups for cAMEs. For convenience, the wrapper function \code{came()} can be used. The current implementation supports RP via the \pkg{rpart} and \pkg{partykit} (CTREE algorithm) packages (classes \code{"PartitioningRPart"} and \code{"PartitioningCTREE"}).
\par
We believe there are two criteria that should guide this process: FME homogeneity within each subgroup and the number of subgroups.
A low number of subgroups is generally preferred. In certain applications, we may want to search for a predefined number of subgroups, akin to the search for a predefined number of clusters in clustering problems. Many RP algorithms do not support searching for a number of subgroups, which is what the \code{"Pruner"} class is intended for.
\item \code{"Pruner"}: To receive a predefined number of subgroups for arbitrary RP algorithms, we follow a two-stage process: grow a large tree by tweaking tree-specific hyperparameters and then prune it back to receive the desired number of subgroups. A \code{"Partitioning"} subclass is implemented such that it can first grow a large tree, e.g., with a low complexity parameter for \pkg{rpart}. Then \code{"Pruner"} iteratively computes the SD of FMEs for each parent node of the current terminal nodes and removes all terminal nodes of the parent with the lowest SD.
\item \code{"PartitioningPlot"}: Decouples visualizations of the separation of $\mathcal{D}$ into subgroups from specific implementations of the \code{"Partitioning"} subclass. Here, we make use of a dependency on \pkg{partykit} for a tree data structure. This allows visualizations of any partitioning with the same methods. The package \CRANpkg{ggparty} \citep{ggparty} creates tree figures that illustrate the partitioning, descriptive statistics for each terminal node, and histograms of FMEs (and optionally NLM values).
\end{itemize}

\section{Conclusion}

This paper introduces the R package \pkg{fmeffects}, the first software implementation of the theory surrounding FMEs. We showcase the package functionality with an applied use case and discuss design choices and implications for future extensions. FMEs are a versatile model-agnostic interpretation method and give us comprehensible model explanations in the form of: if we change $x$ by an amount $h$, what is the change in predicted outcome $\widehat{y}$? FMEs equip stakeholders, including those without ML expertise, with the ability to understand feature effects for any model. We therefore hope that this package will work towards a more widespread adoption of FMEs in practice.
\par
Software development is an ongoing process. As the theory surrounding FMEs evolves, so should the \pkg{fmeffects} package. As noted by \citet{scholbeck_fme}, possible directions for future research include the development of techniques to better quantify extrapolation risk for the selection of step sizes; furthermore, the subgroup search for cAMEs is subject to uncertainties, which may be able to be quantified; and lastly, we may be able to spare computations by searching for representative FMEs, similar to prototype observations that are representative of clusters of observations \citep{tan_data_mining}. Future performance improvements may also be made via parallel computing, which at this point is only implemented for NLM computations.

\newpage
\bibliography{fmeffects}

\newpage
\address{Holger Löwe\\
  Ludwig-Maximilians-Universität in Munich\\
  Germany \\
  \email{hbj.loewe@gmail.com}
  }

\address{Christian A. Scholbeck\\
  Ludwig-Maximilians-Universität in Munich\\
  Munich Center for Machine Learning (MCML)\\
  Germany \\
  \url{https://orcid.org/0000-0001-6607-4895}\\
  \email{christian.scholbeck@stat.uni-muenchen.de}}

\address{Christian Heumann\\
  Ludwig-Maximilians-Universität in Munich\\
  Germany \\
  \email{christian.heumann@stat.uni-muenchen.de}}

\address{Bernd Bischl\\
  Ludwig-Maximilians-Universität in Munich\\
  Munich Center for Machine Learning (MCML)\\
  Germany \\
  \email{bernd.bischl@stat.uni-muenchen.de}}

\address{Giuseppe Casalicchio\\
  Ludwig-Maximilians-Universität in Munich\\
  Munich Center for Machine Learning (MCML)\\
  Germany \\
  \email{giuseppe.casalicchio@stat.uni-muenchen.de}}

\end{article}

\end{document}